\documentclass{ieeeaccess}
\usepackage{cite}
\usepackage{amsmath,amssymb,amsfonts}
\usepackage{algorithmic}
\usepackage{graphicx}
\usepackage{textcomp}
\def\BibTeX{{\rm B\kern-.05em{\sc i\kern-.025em b}\kern-.08em
    T\kern-.1667em\lower.7ex\hbox{E}\kern-.125emX}}

\usepackage[hyphens]{url}  \usepackage{graphicx}    \usepackage{graphicx}  

\usepackage[utf8]{inputenc} \usepackage{lipsum}
\usepackage{url}            \usepackage{booktabs}       \usepackage{amsfonts}       \usepackage{amsmath}
\usepackage{dsfont}
\usepackage{nicefrac}       \usepackage{microtype}      \usepackage{tabularx}
\usepackage{subcaption}
\usepackage[english]{babel}
\usepackage{placeins}
\usepackage{pifont}
\usepackage[rgb,dvipsnames]{xcolor}

\newcommand{\cmark}{\textcolor{OliveGreen}{\ding{51}}}
\newcommand{\xmark}{\textcolor{BrickRed}{\ding{55}}}

\newcommand{\revision}[1]{#1}

\DeclareMathOperator*{\argmax}{arg\,max}

\def\bx{{\mathbf{x}}}
\def\by{{\mathbf{y}}}
\def\bz{{\mathbf{z}}}
\def\bu{{\mathbf{u}}}
\def\R{{\mathbb{R}}}

\newcommand\given[1][]{\:#1\vert\:}

\newcommand{\set}[1]{\{ #1 \}}

\newcommand{\todo}[1]{}

\begin{document}

\history{Date of publication xxxx 00, 0000, date of current version xxxx 00, 0000.}
\doi{XXXXX}

\title{Sparse Gaussian Processes on Discrete Domains}
\author{\uppercase{Vincent Fortuin}\authorrefmark{1}\authorrefmark{*}, \uppercase{Gideon Dresdner}\authorrefmark{1}\authorrefmark{*}, \uppercase{Heiko Strathmann}\authorrefmark{2}, \uppercase{Gunnar R\"atsch}\authorrefmark{1}}
\address[1]{Department of Computer Science, ETH Z\"urich, Z\"urich, Switzerland.}
\address[2]{Gatsby Unit, University College London, London, United Kingdom.}
\address[*]{Equal contribution.}
\tfootnote{VF and GD contributed equally. VF was supported by a PhD fellowship from the Swiss Data Science Center. This work was supported by the grant \#2017-110 of the Strategic Focus Area ``Personalized Health
and Related Technologies (PHRT)'' of the ETH Domain.}

\markboth
{Fortuin \headeretal: Sparse Gaussian Processes on Discrete Domains}
{Fortuin \headeretal: Sparse Gaussian Processes on Discrete Domains}

\corresp{Corresponding author: Vincent Fortuin (e-mail: \texttt{fortuin@inf.ethz.ch}).}

\begin{abstract}
Kernel methods on discrete domains have shown great promise for many challenging data types, for instance, biological sequence data and molecular structure data.
Scalable kernel methods like Support Vector Machines may offer good predictive performances but do not intrinsically provide uncertainty estimates.
In contrast, probabilistic kernel methods like Gaussian Processes offer uncertainty estimates in addition to good predictive performance but fall short in terms of scalability.
While the scalability of Gaussian processes can be improved using sparse inducing point approximations, the selection of these inducing points remains challenging.
We explore different techniques for selecting inducing points on discrete domains, including greedy selection, determinantal point processes, and simulated annealing.
We find that simulated annealing, which can select inducing points that are not in the training set, can perform competitively with support vector machines and full Gaussian processes on synthetic data, as well as on challenging real-world DNA sequence data.
\end{abstract}


\titlepgskip=-15pt

\maketitle

\section{Introduction}\label{introduction}

\begin{figure*}
    \centering
    \includegraphics[width=0.9\textwidth]{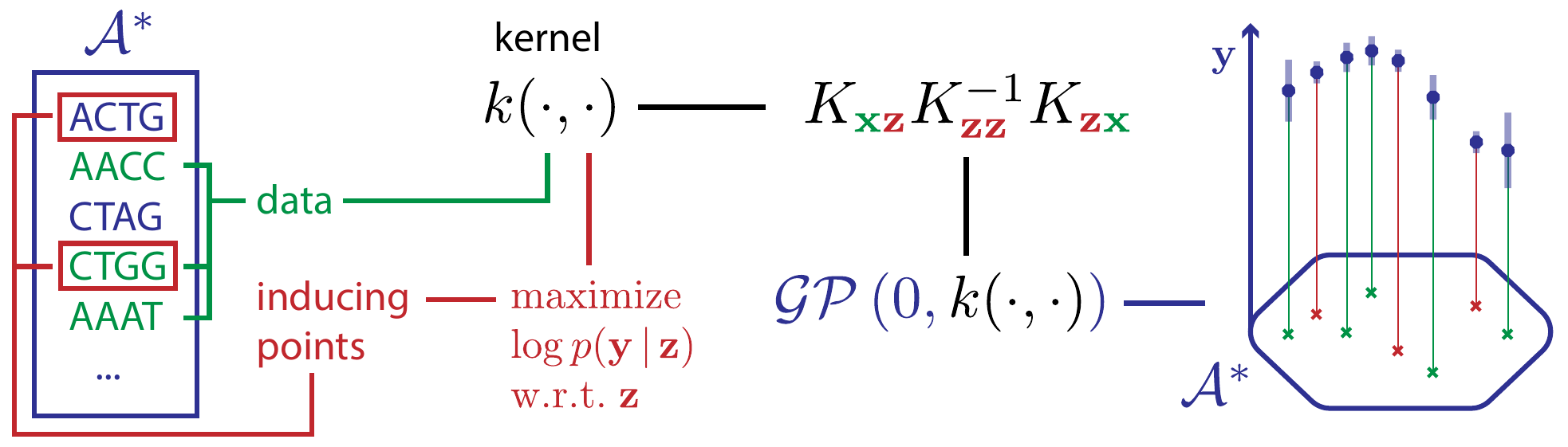}
    \caption{Overview of our proposed framework. The input domain is discrete, in this case the set of strings over some alphabet $\mathcal{A}$. Inducing points for the sparse Gaussian Process are chosen from the data points, but also from the rest of the domain. The choice of inducing points is optimized with respect to the log marginal likelihood. A discrete kernel function is chosen to construct the sparse approximation of the GP's covariance matrix. The GP can then be used to predict latent function values and uncertainties on the input domain.}
    \label{fig:overview}
\end{figure*}

\PARstart{U}{ncertainty} quantification is an increasingly important feature of machine learning models. This is particularly crucial in applications such as in biomedicine \cite{Kononenko2001-sb,ghahramani2015probabilistic,hatch2016snowball}, where prediction errors may have serious repercussions. Consider a wet lab biologist seeking to find a DNA sequence which can be targeted by a drug (for instance, using CRISPR-cas9 \cite{jinek2012programmable}). They have reduced the problem to some number of candidate sequences but to further narrow the selection requires painstaking experiments. If they had a framework that could incorporate their prior knowledge of DNA sequence similarity as well as the results from previous experiments, they could optimally select the best next experiment to perform, thereby saving vast amounts of time and resources. Such a framework would need to perform well under various data sizes as well as provide calibrated uncertainty estimates in order to make an informed decision.

Many problems, like this one, are discrete, involve large datasets, and require well-calibrated uncertainty estimates.
Kernel methods have shown performances that are competitive with deep learning models in such application domains \cite{Morrow2017-gd}, while probabilistic modeling provides a unified framework for prediction and calibrated uncertainty estimates \cite{Murphy2012-uj}.
One class of probabilistic kernel methods that have proven to be useful in various regression and classification settings are Gaussian Processes (GPs) \cite{Rasmussen2006-zv}.
They are data efficient, non-parametric, and have tractable posterior distributions.
Moreover, one can use any kind of likelihood for the generating process, for example, a Bernoulli likelihood in the case of a classification problem.

The main challenge of scaling GPs to large datasets lies in the computational complexity of inference which is cubic in the number of observations. Inducing point methods are the main class of approaches for circumventing this limitation \cite{Quinonero-Candela2005-kt,Titsias2009-dw,Hensman2013-vt,Wilson2015-vx}. These methods aim to use some $m \ll n$ inducing points to reduce the inference complexity to $\mathcal{O}(n m^2)$.

Having reduced the computational complexity of inference, the remaining challenge is to choose the set of inducing points that best approximates the full model \cite{Quinonero-Candela2005-kt}. When the domain is continuous, the locations of the inducing points can be optimized using the gradient of the log marginal likelihood \cite{Snelson2006-wk}. Unfortunately, this gradient-based optimization scheme is not feasible in discrete domains, where the log marginal likelihood is no longer differentiable with respect to the inducing point locations.

In this work, we explore different techniques for choosing inducing points over discrete domains by combining discrete optimization with sparse GP approximations.
In our experiments, we show that our sparse GP framework has comparable performance to full (i.e., not sparsified) GP models as well as Support Vector Machines on biological sequence data.

We make the following contributions:
\begin{itemize}
	\item We present the first empirical assessment of a range of different inducing point selection techniques for Gaussian Processes on discrete domains.
	\item We discuss and evaluate the tradeoffs of these different techniques, for example, in terms of computational complexity and their ability to choose inducing points from outside the training set.
	\item We assess the performance of the models on synthetic data and several challenging real-world datasets.
\end{itemize}
In the following sections, we describe the main components of our framework beginning with sparse GPs, continuing to discrete inducing point selection methods, and concluding with the spectrum string kernel. Each inducing point method corresponds to a different sparse string GP in our framework. For a high level overview of the framework see Figure~\ref{fig:overview}.
Finally, we present experiments comparing each inducing point selection technique in our framework using both Gaussian and Bernoulli likelihoods, that is, binary labels, on synthetic toy data, UCI splicing data \cite{Noordewier1990-tt}, and the DREAM5 dataset \cite{weirauch2013evaluation}.

\section{Sparse Gaussian Process approximations}
\label{sec:sparse_GPs}

Consider a supervised learning problem in which the goal is to estimate a latent function $f:\mathcal X\to\R$ given observed inputs $\bx := (x_1,\ldots,x_n)$ and corresponding outputs $\by := (y_1,\ldots,y_n)$. For the biologist example in the previous section, $f$ could map DNA sequences to drug targetability scores. We assume that our observations are corrupted by additive noise $\eta$, thus $\by = f(\bx) + \eta$, where we have overloaded the notation of the function to be broadcast elementwise. Following a long line of previous work \cite{Rasmussen2006-zv}, we treat the function $f$ as an unobserved random variable with a Gaussian Process prior. Specifically, a GP prior with a zero mean function and a covariance kernel $k(\cdot,\cdot)$:
\[
  f(\cdot)\sim \mathcal{GP}(0, k(\cdot,\cdot))
\]
It follows that the prior on the function outputs, $\mathbf{f}:= f(\bx)$, is given by $\mathcal{N}(0, K_{\bx\bx})$ where $K_{\bx\bx}$ denotes the Gram matrix (also known as the kernel matrix) with $(K_{\bx\bx})_{ij} = k(x_i,x_j)$.

In general, the predictive distribution cannot be solved in closed form but in the special case of Gaussian noise, where $\eta\sim\mathcal{N}(0,\sigma^2)$, the predictive distribution can be computed as
\begin{align*}
	p(\mathbf{f^*} \given \mathbf{y}, \mathbf{x}, \mathbf{x^*}) &= \mathcal{N} \left( \mathbf{m^*}, \mathbf{K^*} \right) \\
	\text{with} \quad \mathbf{m^*} &= K_{*\bx} \left( K_{\bx\bx} + \sigma^2 I \right)^{-1} \mathbf{y} \nonumber \\
	\mathbf{K^*} &= K_{**} - K_{*\bx} \left( K_{\bx\bx} + \sigma^2 I \right)^{-1} K_{\bx*} \; \nonumber
\end{align*}
where the test inputs and outputs are denoted $\bx^*$ and $\mathbf f^*$ respectively and $K_{*\cdot} = K_{\cdot *}^\top$ is shorthand for $K_{\bx^*\cdot}$.

While this closed-form predictive distribution is appealing and has found numerous applications, scaling it to large datasets is fundamentally limited by the matrix inversion $(K_{\bx\bx} + \sigma^2 I)^{-1}$, which requires $\mathcal{O}(n^3)$ operations.

This motivates the use of so-called ``inducing point'' methods which provide a framework for trading model quality for tractability. We assume that there exists a set of $m$ inducing points $(z_1,\ldots,z_m) =: \bz$, $z_i\in\mathcal{X}$ with outputs $\bu:= f(\bz)$ which are distributed $\mathcal{N}(0, K_{\bz\bz})$ according to the prior. Now, we make the modeling assumption that $\mathbf{f}$ and $\mathbf{f^*}$ are conditionally independent given $\mathbf{u}$, that is, $p(\mathbf{f}, \mathbf{f^*}, \mathbf{u}) = p(\mathbf{f} \given \mathbf{u}) \, p(\mathbf{f^*} \given \mathbf{u}) \, p(\mathbf{u})$. We can again solve the inference problem in closed form:
 \begin{align*}
 	p(\mathbf{f^*} \given \mathbf{u}, \mathbf{z}, \mathbf{x^*}) &= \mathcal{N} \left( \mathbf{m^u}, \mathbf{K^u} \right) \\
 	\text{with} \quad \mathbf{m^u} &= K_{*\bz} K_{\bz\bz}^{-1} \mathbf{u} \nonumber \\
 	\mathbf{K^u} &= K_{**} - K_{*\bz} K_{\bz\bz}^{-1} K_{\bz*} \;. \nonumber
 \end{align*}
Note that we have reduced the cubic part of inference from $\mathcal{O}(n^3)$ to $\mathcal{O}(m^3)$, where we can choose $m$, the number of inducing points. Overall, the inference procedure has complexity $\mathcal{O}(n m^2)$ \cite{Quinonero-Candela2005-kt,Titsias2009-dw,Hensman2013-vt,Wilson2015-vx}.

Inducing point methods provide a framework for dramatically decreasing the computational complexity of inference, but we are still left with the problem of choosing the set of inducing points that achieves the best possible approximation with limited resources (namely $m$ inducing points).
This inducing point selection can be cast as an optimization problem in which we are trying to maximize the log marginal likelihood:
\begin{equation}\label{eq:mloglik}
	\log p(\mathbf{y}  \given  \mathbf{z}) = \iint p(\mathbf{y}  \given  \mathbf{f}) \, p(\mathbf{f}  \given  \mathbf{u}) \, p(\mathbf{u}  \given  \mathbf{z}) \, \mathrm{d}\mathbf{u} \, \mathrm{d}\mathbf{f}
\end{equation}
with respect to the locations $\mathbf{z}$.
Standard methods for solving the inducing point selection problem focus on continuous inputs and overlook the case of discrete ones.
In the following section, we tackle this problem on discrete domains using effective and well-tested discrete optimization methods.

\section{Discrete optimization techniques}
\label{sec:discrete_optimization}

The problem we are trying to solve using discrete optimization is $\argmax_\bz \log p(\by\given \bz)$ (Eq.~\ref{eq:mloglik}). In the following sections, we approach this problem using two classical techniques from discrete optimization and one submodular data summarization model: greedy selection, simulated annealing, and determinantal point processes.

\subsection{Greedy selection}

Greedy inducing point selection dates back to early works on sparse Gaussian Processes \cite{Csato2001-ro,Seeger2003-zu,Titsias2009-dw}. The algorithm is initialized with an empty set of inducing points and at each iteration greedily selects the next observation in the data that maximizes the marginal likelihood $p(\by \given \bz)$ (Eq.~\ref{eq:mloglik}). Thus, the set of inducing points is a mere subset of the original data. This approach is justified by the fact that the marginal likelihood is strictly monotonic in the number of inducing points. Hence, adding a new inducing point is always guaranteed to increase the objective.

This technique is conceptually simple and easy to implement, but it comes with the major drawback that the inducing points can only be selected from the training set.
Especially in high-dimensional discrete spaces, the training set might only span a small fraction of the total space, so this can be a strong limitation.

Natural extensions of this method include selecting several inducing points instead of just one at every iteration, swapping inducing points between the training set and the inducing point set \cite{cao2013efficient}, and optimizing a variational lower bound on the likelihood rather than the likelihood itself \cite{Titsias2009-dw}.
Since we mainly include this method as a baseline in our experiments, we leave the exploration of these extensions to future work.

\subsection{Simulated annealing}

Simulated annealing is a sampling-based approach which starts with an initial guess $S_0 := \set{\bz_1,\ldots,\bz_m}$ and a loss function $\mathcal L(\cdot)$ to be optimized. At each iteration the algorithm perturbs an element of the set and decides whether or not to accept this new perturbation as the next state. To make this decision, an energy term is computed from the current iterate $S_{t-1}$ and the proposal $\hat S$, $E_t = \mathcal L(S_{t-1}) - \mathcal L(\hat S)$. The new proposal is then accepted with probability
\[
	P_{\text{accept}} (\hat S) = \min \left( 1, \exp \left( \frac{E_t}{T_t} \right) \right) \; ,
\]
where $T_t$ is known as the temperature parameter and is usually chosen with an exponential decay rate in $t$.

Since we are working on discrete string domains, we define a perturbation to be a change of one or more characters in a given string.
Determining the number of characters to change requires careful fine tuning.
In our experiments we chose the most conservative setting of perturbing just a single character at each step.
The loss function is again naturally defined as $\mathcal L(\mathbf{z}) := \log p(\mathbf{y} \given \mathbf{z})$.

Crucially, this perturbation approach allows the simulated annealing to explore inducing points from the entire input space, and not only from the training set.
This makes it more flexible than greedy selection or the determinantal point process described in the following.

\subsection{Determinantal Point Processes}

A Determinantal Point Process (DPP) with kernel $k$ is a distribution over subsets of observations \cite{Kulesza2012Jul}.
Given a subset of points $\bz \subseteq \bx$ with $| \bz | = m$ as above, its probability is defined as\footnote{Strictly speaking, this is a so-called ``L-ensemble,'' not a DPP \cite{Kulesza2012Jul}. Furthermore, DPPs can be defined more generally over uncountable sets but for the sake of clarity, we focus on what is relevant for our work. This also agrees with the majority of DPP applications in machine learning.}
\begin{equation}
  P(\bz) = \frac{\det(K_{\bz\bz})}{\det(K_{\bx\bx} + I)}
\end{equation}
where $I$ is the identity matrix. Intuitively, the determinant of the Gram matrix $K_{\bz\bz}$ represents the volume of the parallelepiped spanned by the features in $\bz$. Therefore, subsets of high probability have a large volume in feature space which in turn implies diversity. In their recent work, \cite{Burt2019Mar} showed analytically that with $\mathcal{O}(\log N)$ inducing points sampled from a DPP, the sparse GP is close to the full GP in KL-divergence. While this result only holds for squared-exponential kernels, we consider it to be a theoretical motivation for using DPPs for inducing point sampling.

While the normalization constant, $\det(K_{\bx\bx} + I)$, is notably available in closed form, this is not relevant for our purposes since it still requires $\mathcal O(n^3)$ operations to compute. Instead, we use fast MCMC-based sampling methods for our DPPs \cite{Li2016Mar}.
Note that this could in principle be extended to sampling from the whole input space, but it would make the normalization constant intractable and require further approximations.
For simplicity, we thus resort to just sampling inducing points from the training set in this work (similar to the greedy approach above).

\section{String kernels}
\label{sec:string_kernels}

While our framework is fully general, in this work we focus our experiments on biological sequences, which are an important real-world discrete data domain. One key aspect of many biological sequences is translation invariance. In this section, we describe $n$-gram-based string kernels which are designed to exhibit this property and thus explicitly incorporate our biological prior knowledge. In cases where full translation-invariance is not a desired property, a practitioner can choose from the vast literature on kernel methods to select a more appropriate prior for the GP.

Specifically, in our work we use the spectrum kernel \cite{Leslie2002-zj}, which was designed for protein sequences and has also been successfully applied to other types of biological sequences \cite{Ben-Hur2008-wo}. There are existing applications which use string kernels in Gaussian Processes, but they use small datasets ($n\approx 280$) where full Gaussian Process inference is viable \cite{Stegle2009-sn}.
In this work, we enable the extension of these methods to larger data sets, through the use of sparse GP approximations.

Given an alphabet $\mathcal A$, we denote the input domain of all strings of finite length as $\mathcal X = \mathcal A^*$. The $n$-th order spectrum kernel is defined over this domain as
\begin{align*}
	k_n(x,x') &= \langle \Phi_n(x), \Phi_n(x') \rangle \\
	\text{with} \quad \Phi_n(x) &= \left[ \phi_a(x) \right]_{a \in \mathcal{A}^n} \nonumber
\end{align*}
where $\phi_a(x)$ is the number of times that the string $a\in\mathcal A^n$ appears as a substring in $x$. This is essentially a bag-of-$n$-grams model.

While the set $\mathcal{A}^k$ might be prohibitively large, thus making the feature maps $\Phi_n(x)$ prohibitively high dimensional, it can easily be seen that we can compute $k_n(\cdot,\cdot)$ without having to represent $\Phi_n(x)$ explicitly. For two strings of arbitrary length, $x \in \mathcal{A}^{l(x)}$ and $x' \in \mathcal{A}^{l(x')}$, the kernel can be rewritten as
\[
	k_n (x, x') = \sum_{i=0}^{l(x) - n} \sum_{j=0}^{l(x') - n} \mathds{1} \lbrack x_{i:i+k} = x'_{j:j+k} \rbrack
\]
Computing this kernel na\"ively has complexity $\mathcal{O}(l^2)$ where without loss of generality $l := l(x) \geq l(x')$. This can be further improved using suffix trees, resulting in a complexity of $\mathcal{O}(kl)$ with $k < l$ \cite{Leslie2002-zj}.

These three components---a GP prior represented by the choice of the kernel function, an inducing point GP approximation, and finally a method for selecting inducing points from a discrete input space---provide a unified framework for supervised learning over discrete input spaces using GPs. This framework not only has good predictive performance but also provides superior uncertainty estimates which we demonstrate in the following experiments.
The main challenge within this framework remains the selection of inducing points from the discrete domains, which is why we will thoroughly assess the different techniques and their tradeoffs in the following. 

\section{Experiments}\label{experiments}

\begin{table*}
\centering
\caption{Performance comparison of different inducing point optimization methods and a full GP on a toy data regression task. Means and their standard errors are computed over 100 runs of the experiment. Note that the full GP is included as a gold standard, but is not considered in the actual comparison because it is not a scalable method.}
        \label{tab:toy_data_regression}

\begin{tabular}{lrrrr}
\toprule
Method &           Likelihood &       Val. likelihood &              Val. MSE &                 Training time [s] \\
\midrule
full                       &   -2861.02 $\pm$ 0.00 &  -1368.65 $\pm$ 0.00 &      7.30 $\pm$ 0.00 &         3.15 $\pm$ 0.72 \\
\midrule
SA                         &  -1847.83 $\pm$ 6.43 &  -1879.09 $\pm$ 4.92 &      3.96 $\pm$ 0.16 &    1220.00 $\pm$ 272.10 \\
greedy\_10percent           &  -1776.35 $\pm$ 2.48 &  -1818.26 $\pm$ 2.42 &      3.54 $\pm$ 0.11 &     695.45 $\pm$ 155.33 \\
greedy\_25percent           &  -1773.27 $\pm$ 2.11 &  -1815.28 $\pm$ 2.12 &      3.49 $\pm$ 0.10 &    2244.93 $\pm$ 541.05 \\
greedy\_50percent           &  -1770.75 $\pm$ 1.85 &  -1813.13 $\pm$ 1.87 &      3.52 $\pm$ 0.08 &   5884.47 $\pm$ 1401.49 \\
greedy\_100percent          &  -1768.75 $\pm$ 1.64 &  -1811.09 $\pm$ 1.76 &      3.50 $\pm$ 0.09 &  17882.54 $\pm$ 3256.66 \\
dpp\_10percent &  -1841.38 $\pm$ 7.79 &  -1874.17 $\pm$ 6.20 &      4.20 $\pm$ 0.19 &        13.44 $\pm$ 2.52 \\
dpp\_50percent  &  -1839.63 $\pm$ 7.52 &  -1873.01 $\pm$ 6.26 &      4.23 $\pm$ 0.24 &        39.03 $\pm$ 6.26 \\
dpp\_100percent     &  -1841.28 $\pm$ 7.65 &  -1874.11 $\pm$ 6.33 &      4.20 $\pm$ 0.22 &       71.20 $\pm$ 11.45 \\
random                     &  -1849.43 $\pm$ 8.29 &  -1882.46 $\pm$ 6.78 &      4.52 $\pm$ 0.28 &         6.77 $\pm$ 1.57 \\
\bottomrule
\end{tabular}
\end{table*}

\begin{table*}
\centering
\caption{Performance comparison of different inducing point optimization methods, a full GP and an SVM on a toy data classification task. Means and their standard errors are computed over 100 runs of the experiment. Note that the full GP is included as a gold standard, but is not considered in the actual comparison because it is not a scalable method.}
        \label{tab:toy_data_classification}

\begin{tabular}{lrrrr}
\toprule
Method &   Val. likelihood &              Val. AUPRC &                 Training time [s] \\
\midrule
full & -432.65 $\pm$ 0.00 & 0.439 $\pm$ 0.000 & 2.06 $\pm$ 0.01 \\
SVM & - & 0.420 $\pm$ 0.000 & 0.54 $\pm$ 0.00\\
\midrule
SA & -345.82 $\pm$ 0.06 & 0.438 $\pm$ 0.000 & 2619.56 $\pm$ 16.87  \\
greedy\_10percent & -346.20 $\pm$ 0.02 & 0.397 $\pm$ 0.001 & 92.99 $\pm$ 1.02  \\
dpp\_100percent & -346.38 $\pm$ 0.01 & 0.395 $\pm$ 0.001 & 2.55 $\pm$ 0.02 \\
random  & -346.42 $\pm$ 0.01 & 0.392 $\pm$ 0.001 & 0.94 $\pm$ 0.01  \\
\bottomrule
\end{tabular}
\end{table*}

\begin{figure*}
\centering
\begin{subfigure}[t]{0.45\textwidth}
\centering
\includegraphics[scale=0.5]{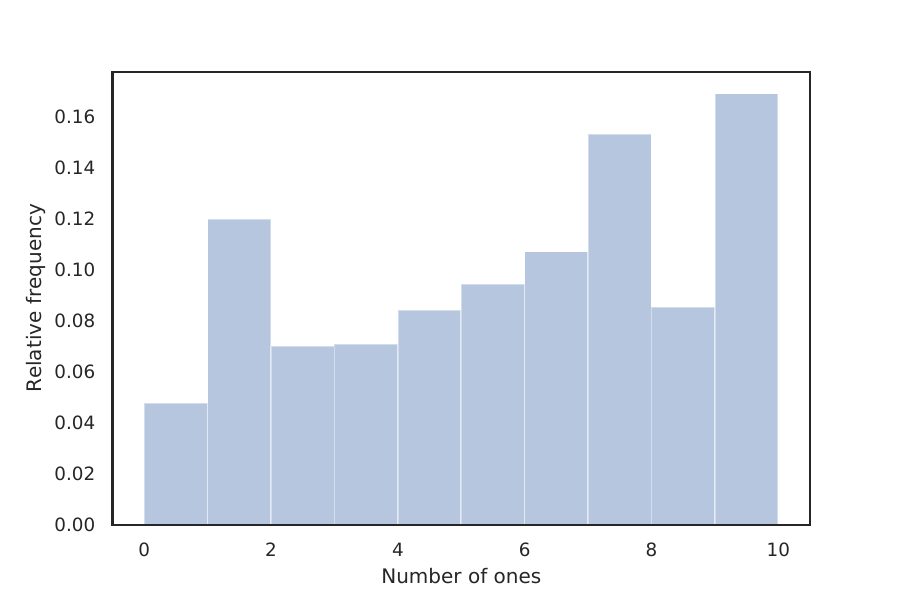}
\subcaption{Regression}
\end{subfigure}
\begin{subfigure}[t]{0.45\textwidth}
\centering
\includegraphics[scale=0.5]{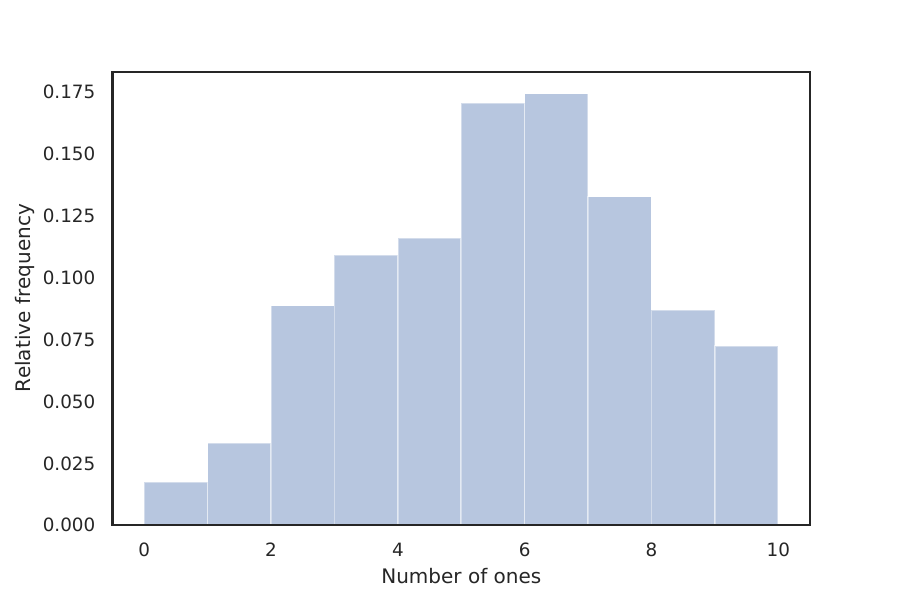}
\subcaption{Classification}
\end{subfigure}
\caption{Visualization of the inducing points that are chosen by the sparse GP optimization. The histograms show the distribution of the number of counted elements in the inducing point strings. It can be seen that the inducing points are distributed evenly over the range of all strings in the regression setting (a), while they accumulate close to the decision boundary between 5 and 6 in the classification task (b).
}
\label{fig:inducing_points}
\end{figure*}

We compared different inducing point optimization methods for sparse GPs on toy datasets in regression and classification settings. We then validated our framework's performance on two real-world DNA sequence datasets from the UCI repository \cite{Dheeru2017-ap} and the DREAM5 dataset \cite{weirauch2013evaluation}. We used support vector machines (SVMs) \cite{hearst1998support} with \emph{post hoc} uncertainty calibration \cite{Platt1999-zy} as a competitive benchmark method for the classification tasks.

We find that our sparse string GP framework performs well when compared to full string GPs in these diverse settings.
Moreover, the inducing points selected by the algorithm align well with the natural intuition for inducing points on continuous domains. Our framework offers comparable predictive performance with SVMs but yields superior uncertainty calibration.

\subsection{\revision{Implementation}}

\revision{If not otherwise noted, all GPs and SVMs use a spectrum kernel as implemented in \texttt{Shogun} \cite{Sonnenburg2010-zp,Sonnenburg2017-el}.
For fitting the GPs, we used the \texttt{GPy} framework \cite{Hensman2013-vt}.
For fitting the SVMs, we used the \texttt{sklearn} package \cite{Pedregosa2011-yd}.}

\revision{Note that in the regression experiments, the GP posterior can be computed in closed form (as described in Sec~\ref{sec:sparse_GPs}), since the Gaussian likelihood is conjugate with the Gaussian prior.
However, in the classification examples, we need to use a non-conjugate Bernoulli likelihood, and thus have to resort to approximate inference.
In our experiments, we use expectation propagation for the approximate inference \cite{Minka2001-nf, Rasmussen2006-zv}, since it is fast and yields good performance.
We use it with the \texttt{GPy} standard parameters.
Alternatively, one could use Markov Chain Monte Carlo (MCMC) inference to get an even better (that is, asymptotically exact) approximation to the posterior.
However, this would be computationally much more expensive and would likely outweigh all the computational benefits of our sparse approximation, which is why we have not tried this approach.
}

\revision{Since the SVM does not natively output probabilities, we have to calibrate it in order to turn the SVM predictions into probabilities.
The \emph{Calibrated SVM} uses a technique called \emph{Platt scaling} \cite{Platt1999-zy}.
It performs a logistic regression on the SVM outputs and calibrates it using a cross-validation on the training data.
}

\subsection{Performance evaluation}

In order to assess the predictive performance of our GP models on regression and classification tasks, we use the mean squared error (MSE) and area under the precision-recall curve (AUPRC), respectively.
We use calibration curves \cite{Zadrozny2002-pa,guo2017calibration} to assess uncertainty calibration.
Moreover, we report the mean absolute deviation (AD) of the calibration curves from the diagonal.
Note that a perfectly calibrated classifier would lie directly on the diagonal of the plot and hence yield an AD of zero.

\subsection{Inducing point optimization for regression and classification}

We developed a simple toy experiment as a controlled setting for our initial comparisons.
It is composed of 1000 strings of a 4 character alphabet~(inspired by the DNA bases,~`A',`C',`T',`G')~each of which has length 100.
We first generated a set of 100 strings of length 5 which we call the \emph{library}.
Each example in the toy dataset contains some copies of a particular element of the library, distributed uniformly at random through the sequence.
The other characters are selected uniformly at random from the alphabet.
The label for each example is the number of elements of the particular library sequence it contains.
We think of this as a discrete dataset with 100 clusters of strings, corresponding to the elements of the library.

This toy dataset challenges the inducing point methods to effectively summarize the data for the prediction task. We use 100 inducing points for all the experiments.
While in general one should perform Bayesian model selection via log marginal likelihood \cite{Rasmussen2006-zv}, to select the optimal kernel hyperparameter $k$, in this case we know the optimal value should be $k=5$, since this is the size of the strings in the library.

In Tables~\ref{tab:toy_data_regression} and \ref{tab:toy_data_classification}, we compare a full GP model, sparse GPs with different inducing point selection methods, and SVMs. For the inducing point methods we used randomly chosen inducing points (\emph{random}), greedy selection (\emph{greedy}), simulated annealing (\emph{SA}), and DPP (\emph{DPP}) methods. For the greedy selection, we experimented with performing it on random subsets to improve performance. Thus, \emph{greedy\_10percent} corresponds to performing a greedy selection on a uniformly random selection of 10\% of the training data. For the DPP, we experimented with different MCMC steps. Thus \emph{dpp\_10percent} corresponds to taking 10\% of recommended $nm$ steps until complete mixing.

Our results confirm the intuition that the sparse GP approximations cannot match the performance of the full GP, neither with respect to log-likelihood nor with respect to predictive performance.
However, our results also demonstrate that inducing point selection is crucial in improving the performance of the sparse GPs. With careful selection of inducing points---either greedily or via simulated annealing---the sparse models can approach the performance of the full model.

Moreover, there is a clear tradeoff between runtime and performance of the methods.
The methods with the longest runtime, particularly the simulated annealing, achieve the best results among the sparse GP models, while for instance the DPP model offers a much more attractive runtime, but slightly lower performance.
It depends therefore on the practitioner's choice whether runtime or performance is a more important selection criterion.
This also suggests that the ability of the simulated annealing to select inducing points from outside the training set can be beneficial in this application.

The inducing points chosen in both regression and classification settings follow a natural intuition. In the regression task, the model has to count the number of library elements equally well across all parts of the space. In the classification task, a more precise count close to the decision boundary is crucial for minimizing classification errors. Figure~\ref{fig:inducing_points} clearly demonstrates this behavior.

This experiment shows that our sparse GP framework approaches the predictive performance of a full GP, while also outperforming baseline SVM methods.
We also find that inducing point selection in discrete string space follows our general intuition for inducing point selection in continuous spaces.

\begin{figure}
\centering
\includegraphics[width=\columnwidth]{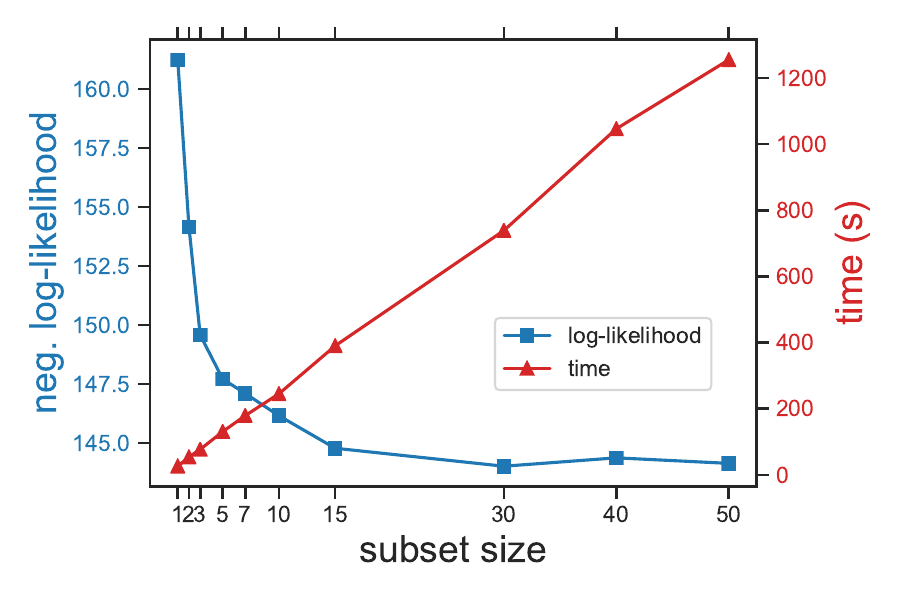}

\caption{Test log-likelihood and runtime as a function of subset size in greedy subset selection. The values are means over 100 runs of the experiment, selecting 4 inducing points each.
}
\label{fig:greedy_subset}
\end{figure}

\begin{table}[t]
    \centering
    \caption{Comparison of training and inference time complexity for full GPs and the different sparse GPs. $n$ is the number of training points, $m$ the number of inducing points, $s$ is the subset size for the greedy subset selection, and $k$ the number of iterations for the simulated annealing.}
    \begin{tabular}{lll}
        \toprule
        Method & Training time & Inference time \\
         \midrule
         full GP & $\mathcal{O}(1)$ & $\mathcal{O}(n^3)$\\
         random & $\mathcal{O}(m)$ & $\mathcal{O}(n m^2)$\\
         greedy & $\mathcal{O}(n^2 m^3)$ & $\mathcal{O}(n m^2)$\\
         greedy subset & $\mathcal{O}(s n m^3)$ & $\mathcal{O}(n m^2)$\\
         SA  & $\mathcal{O}(k n m^2)$ & $\mathcal{O}(n m^2)$ \\
         DPP  & $\mathcal{O}(nm)$ & $\mathcal{O}(n m^2)$ \\
         \bottomrule
    \end{tabular}
    \label{tab:time_complexities}
\end{table}

There is a natural tradeoff between the fast but suboptimal performance of random inducing point selection and the slow but superior performance of greedy selection. We explored this tradeoff by restricting the greedy inducing point selection over the entire dataset to a randomly sampled subset of the data. We then varied the size of this random subset. In the case when the random subset is the same size as the original dataset, one recovers the greedy algorithm. See Table~\ref{tab:time_complexities} for a summary of the time complexities of all methods discussed.

The results are depicted in Figure~\ref{fig:greedy_subset}.
There is a clear tradeoff between performance and runtime, both of which increase as subset size increases.
However, the runtime grows linearly with the subset size (as expected), while the likelihood converges.
We expect that finding the optimal subset size will highly depend on the application both in terms of specific properties of the dataset as well as the computational resources available to the practitioner.

\subsection{Real world DNA sequence data}

\begin{figure}
\centering
\includegraphics[scale=0.42]{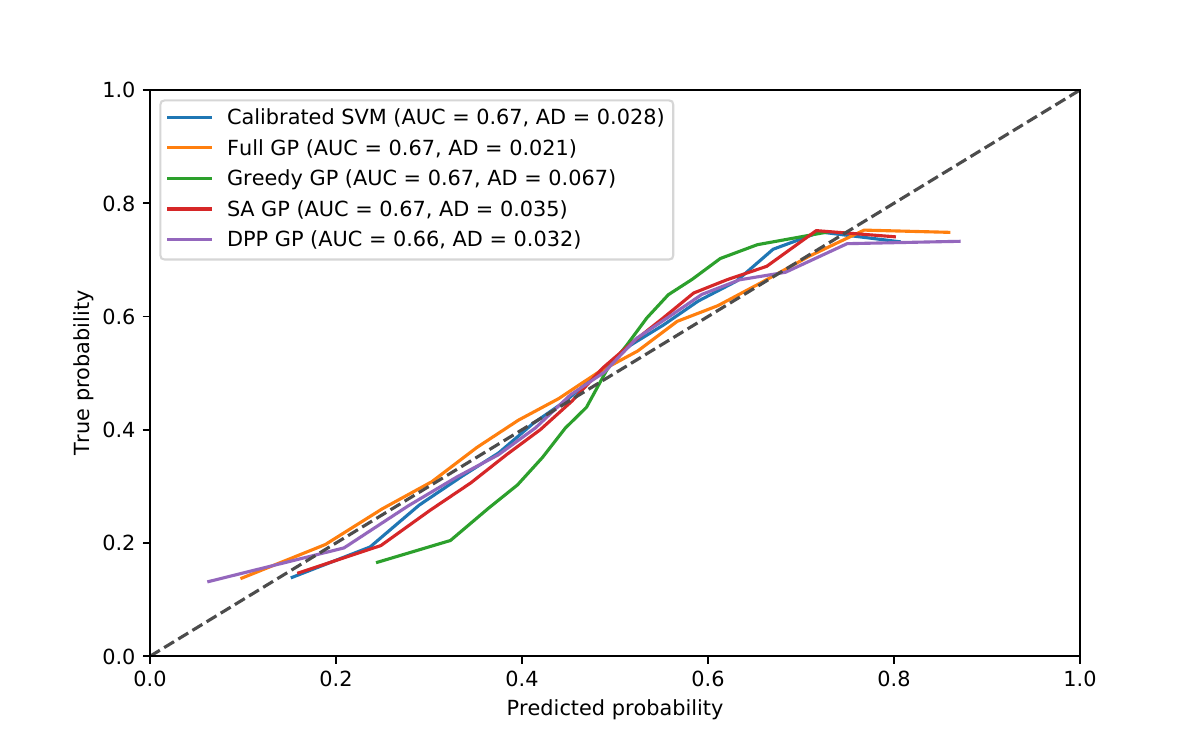}
\label{fig:UCI_splicing_calibration}
\caption{Calibration curves for calibrated SVM, full GP and sparse GP predictions on the UCI DNA splicing sequence datasets. Areas under the precision-recall curve (AUPRC) as well as mean absolute deviation (AD) are reported for all methods in the legend. It can be seen that all methods achieve comparable performances in terms of predictive accuracy and calibration.
}
\label{fig:UCI_data_calibration}
\end{figure}

To validate our models on real-world data, we performed classification on the \emph{UCI splicing} dataset \cite{Noordewier1990-tt}.
Moreover, to demonstrate the scalability of these methods, we performed regression on the DREAM5 dataset \cite{weirauch2013evaluation}.
As with our experiments on synthetic data, we aim to compare the predictive performance and uncertainty calibration against SVMs for classification.
The splicing dataset contains 3,190 sequences of 60 nucleotides each which have to be classified into splicing and non-splicing sites. We applied our methods to the pTH2427 transcription factor of the DREAM5 dataset which contains a total of 32,896 short sequences of length~8. On the~DREAM5 dataset, we performed a 70-30\% train-test split.

\begin{table*}
  \centering
  \caption{Comparison between greedy and DPP methods on the DREAM5 dataset. The kernel matrix for the DPP was normalized as $k(x,y) / \sqrt{k(x,x)k(y,y)}$. The greedy method was run on uniform random subsets of size 20 of the training data.}
  \begin{tabular}{lrrrr}
    \toprule
    Method &            Train likelihood &        Val. likelihoood &              Val. MSE &                Inference time [s] \\
    \midrule
    greedy &  -29702.52 $\pm$ 0.00 &  -14637.15 $\pm$ 0.00 &       0.87 $\pm$ 0.00 &   590.16 $\pm$ 122.43 \\
    dpp &   -29702.52 $\pm$ 0.00 &   -14637.15 $\pm$ 0.00 &       0.87 $\pm$ 0.00 &    1943.38 $\pm$ 291.71 \\
    random                     &  -29723.81 $\pm$ 76.48 &  -14645.75 $\pm$ 34.01 &       0.87 $\pm$ 0.00 &      305.83 $\pm$ 18.47 \\
    \bottomrule
  \end{tabular}
  \label{tab:dream}
\end{table*}

\begin{table*}
	\caption{\revision{Comparison of our approach and related models in terms of different desirable properties.}}
	\centering
	\begin{tabular}{lcccc}
        \toprule
        Method & Uncertainty & Scalable & Discrete domain & Inducing points from full domain \\
         \midrule
         \revision{SVM \cite{hearst1998support}} & \xmark & \cmark & \cmark & \xmark \\
         \revision{Full GP \cite{Stegle2009-sn, moss2020boss}} & \cmark & \xmark & \cmark & \xmark \\
         \revision{Variational GP \cite{Titsias2009-nm}} & \cmark & \cmark & \xmark & \cmark \\
         \revision{Greedy GP \cite{cao2013efficient}} & \cmark & \cmark & \cmark & \xmark \\
         \revision{DPP-GP \cite{Burt2019Mar}} & \cmark & \cmark & \cmark & \xmark \\
         \revision{SA-GP (ours)} & \cmark & \cmark & \cmark & \cmark \\
         \bottomrule
    \end{tabular}
    \label{tab:related_work}
\end{table*}

\begin{table}
    \centering
    \caption{Comparison of inference times and predictive performances of full GPs, our sparse simulated annealing GPs and SVMs on a subset of the UCI splicing data. Values are means and their standard errors of 100 runs.}
    \begin{tabular}{lrr}
        \toprule
        Method & AUPRC & Inference time [s] \\
         \midrule
         full GP & 0.678 $\pm$ 0.002 & 669.8 $\pm$ 11.1 \\
         Sparse GP & 0.676 $\pm$ 0.002 & 15.0 $\pm$ \enspace 0.2 \\
         SVM & 0.674 $\pm$ 0.002 & 7.7 $\pm$ \enspace 0.1 \\
         \bottomrule
    \end{tabular}
    \label{tab:inference_times}
\end{table}

We compared a kernel SVM against a full GP and our sparse GPs with inducing points selected greedily and by simulated annealing.
Note that the splicing data as well as the DREAM5 data are too large for feasible full GP inference.
To provide a fair comparison, we report inference times of our sparse GP in comparison with the full GP on a randomly selected subset of the splicing data in Table~\ref{tab:inference_times}.
It can be seen that our sparse GP speeds up inference by more than one order of magnitude while still yielding comparable predictive performance (c.f.\ Tab.~\ref{tab:time_complexities}).

The sparse GPs use 50 inducing points on the splicing data.
The order of the spectrum kernel was chosen to be $k=3$ by all GPs through log marginal likelihood optimization.
The performance of the methods in terms of area under the precision-recall curve (\emph{AUPRC}) and calibration is measured by a 2000/1190 train-test-split on the splicing data.
Results are reported in Figure~\ref{fig:UCI_data_calibration}.

For the DREAM5 dataset (Tab.~\ref{tab:dream}) we used a kernel of size $k=3$. This was motivated both by biological considerations of the size of a codon and also by the fact that the sequences are short---only 8 characters. We found that when set to the right random subset size (in this case 20), greedy selection could outperform the DPP in both runtime and performance.

If we compare the calibration of the different methods on the classification task, it can be seen that the various sparse GP models, and the calibrated SVM are comparable in terms of calibration and predictive performance. (Fig.~\ref{fig:UCI_data_calibration}).
The calibration ranking among the GPs is analogous to the one for the log-likelihoods, that is, the sparse GP with inducing points optimized by simulated annealing ranks second, the one with greedily selected points third.

These experiments show that our framework yields a comparable performance with full GP inference as well as kernel SVMs on real world DNA sequence classification tasks.
Moreover, it scales to larger datasets where full GP inference is computationally infeasible.
It also shows that greedy selection can perform better than the theoretically motivated DPP sampling, while simulated annealing generally performs best, possibly due to its more flexible inducing point selection from outside the training set.

\section{Related work}

\paragraph{Sparse Gaussian processes}
This work builds upon the rich literature on inducing point methods for Gaussian Processes (see \cite{Quinonero-Candela2005-kt} and references therein).
Recent work in this domain has utilized variational approximations \cite{Titsias2009-nm} and certain geometrical structures \cite{Wilson2015-vx}.
\revision{Furthermore, it has been proposed to use spherical harmonic features \cite{dutordoir2020sparse}, orthogonal inducing points \cite{shi2020sparse}, and doubly sparse GPs \cite{adam2020doubly}.}
Unfortunately, all these advances are limited to continuous input spaces, so we are forced to resort to more conventional inducing point methods in this work.

\paragraph{Gaussian processes on discrete domains}
Many kernels have been devised to work well on discrete domains, for instance, on strings \cite{Leslie2002-zj} or graphs \cite{Kondor2002-em}.
These have been used successfully in combination with SVMs or similar linear models for problems in biology \cite{Ben-Hur2008-wo,Morrow2017-gd}, chemistry \cite{Mahe2005-el,Mahe2009-zj}, and natural language processing \cite{Lodhi2002-hc}.
Using discrete kernels in GPs is a relatively unexplored area, possibly due to the difficulties in hyper-parameter optimization and inducing point selection. Discrete kernels have been used on graphs \cite{Venkitaraman2018-ov} and strings \cite{Beck2017-ae} (also for biological problems \cite{Stegle2009-sn}), but so far only on relatively small datasets with full GPs.
In parallel work, \cite{moss2020boss} study GPs on strings for Bayesian optimization, but their problems are small they do not use inducing points or any other scalability approach.

\paragraph{Discrete inducing point selection}
While the greedy inducing point selection approach has already been proposed in early work on GP regression \cite{Smola2001-ym, lawrence2003fast, Seeger2003-zu}, these works have not particularly assessed its performance on discrete GPs.
To the best of our knowledge, the only work that previously studied discrete sparse GPs are \cite{cao2013efficient}, who also use a greedy approach, although with greedy swapping between the inducing point set and the training set, instead of greedy forward selection.
We are the first to additionally study DPPs for discrete inducing point selection, as well as simulated annealing, thus enabling the use of inducing points that are not included in the training set.
\revision{For an overview comparison of our proposed simulated annealing approach with the other related models in terms of uncertainty estimation, scalability, applicability to discrete domains, and selection of the inducing points, see Table~\ref{tab:related_work}.}

\section{Conclusion and future work}

In this work, we explore different inducing point selection techniques for sparse Gaussian Processes on discrete domains.
We found empirically that our proposed method using simulated annealing gives the best overall predictive performance and uncertainty estimates.
This method also yields favorable runtimes on larger datasets and more non-standard likelihoods.

Moreover, we showed that our models perform competitively with SVMs on toy data as well as real-world DNA sequence data in terms of predictive performance, while offering better calibrated uncertainty estimates in some settings.

There are many directions for future work.
First, developing a closer integration between discrete optimization and the marginal likelihood of the GP would improve both the approximation quality as well as the runtime of the inducing point algorithm.
An orthogonal direction is a fully Bayesian treatment of the string kernel hyperparameter $k$, namely treating $k$ as a random variable with a prior and performing inference on it.
Finally, we would like to see existing GP software packages extend their abstractions to discrete kernels with hyperparameters that are not differentiable.

In conclusion, we would advise practitioners to use our framework on discrete problems where datasets are too large for full Gaussian Process models but uncertainty estimates are still desirable. Furthermore, in cases where likelihoods other than Gaussian or Bernoulli are required, standard regression and classification techniques are inapplicable, whereas our framework provides a principled and flexible solution.

\bibliographystyle{IEEEtran}
\bibliography{library,gideon}

\newpage

\begin{IEEEbiography}
[{\includegraphics[width=1in,height=1.25in,clip,keepaspectratio]{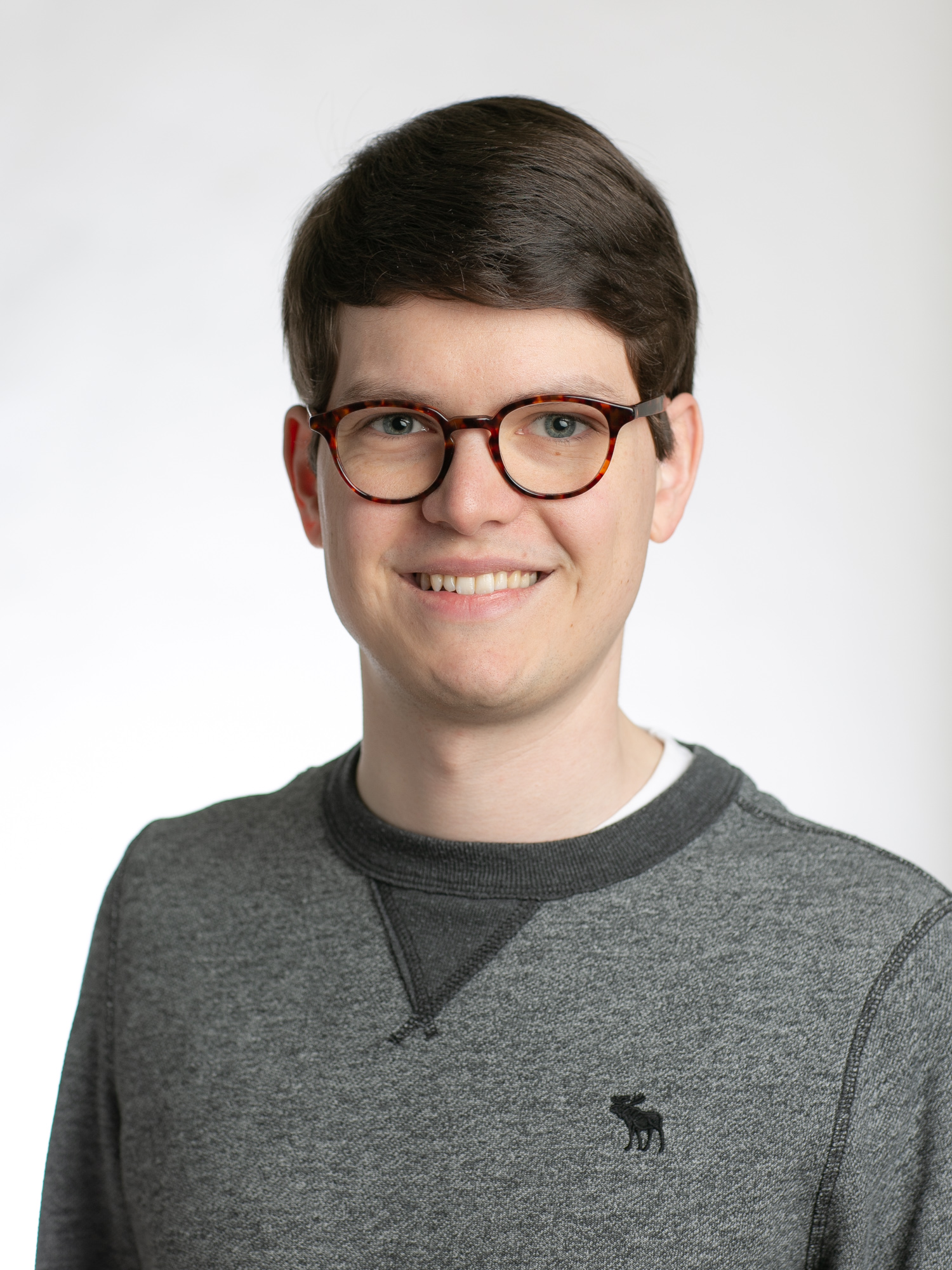}}]{Vincent Fortuin} is a doctoral student at ETH Z\"urich where he works on Bayesian deep learning, representation learning, and approximate inference.
\end{IEEEbiography}

\begin{IEEEbiography}
[{\includegraphics[width=1in,height=1.25in,clip,keepaspectratio]{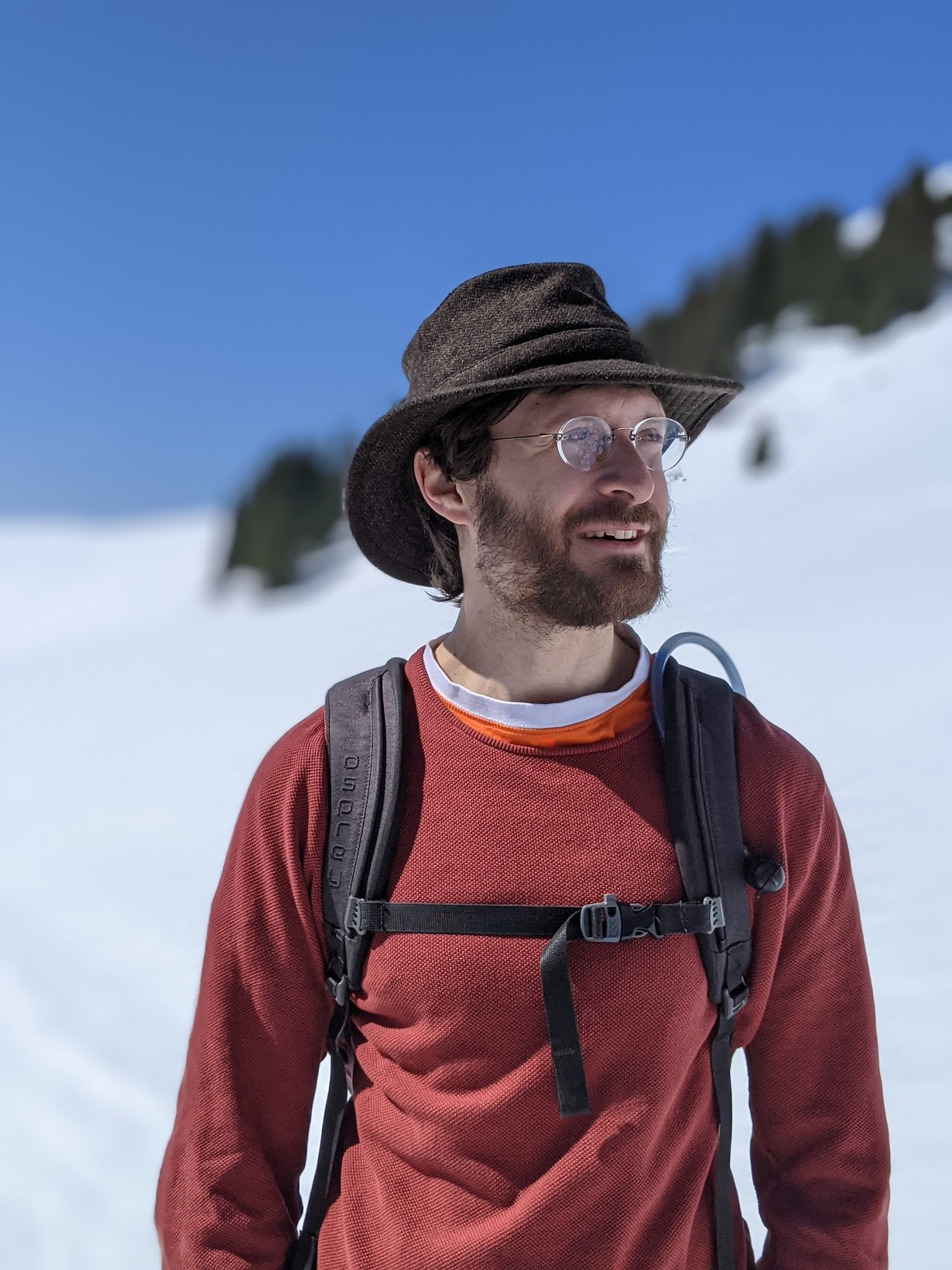}}]{Gideon Dresdner} is a doctoral student at ETH Z\"urich where he works on convex optimization and Bayesian inference.
\end{IEEEbiography}

\begin{IEEEbiography}
[{\includegraphics[width=1in,height=1.25in,clip,keepaspectratio]{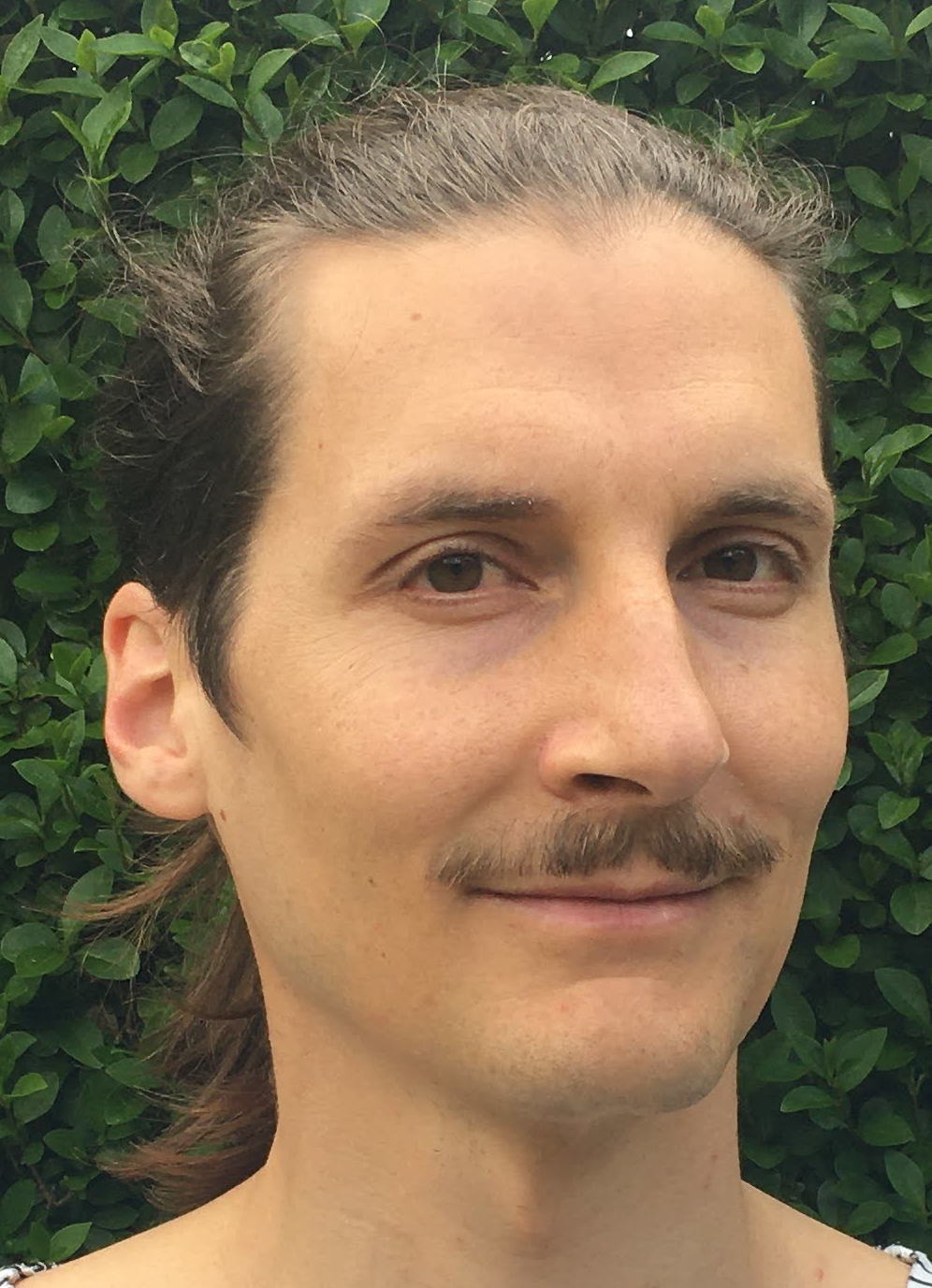}}]{Heiko Strathmann} is a Research Scientist at Deepmind. He completed his PhD at the Gatsby Unit, UCL, and since spent time at the ETHZ, CH and The Alan Turing Institute, UK. His research interests include generative models, kernel methods, and Bayesian inference.
\end{IEEEbiography}

\begin{IEEEbiography}
[{\includegraphics[width=1in,height=1.25in,clip,keepaspectratio]{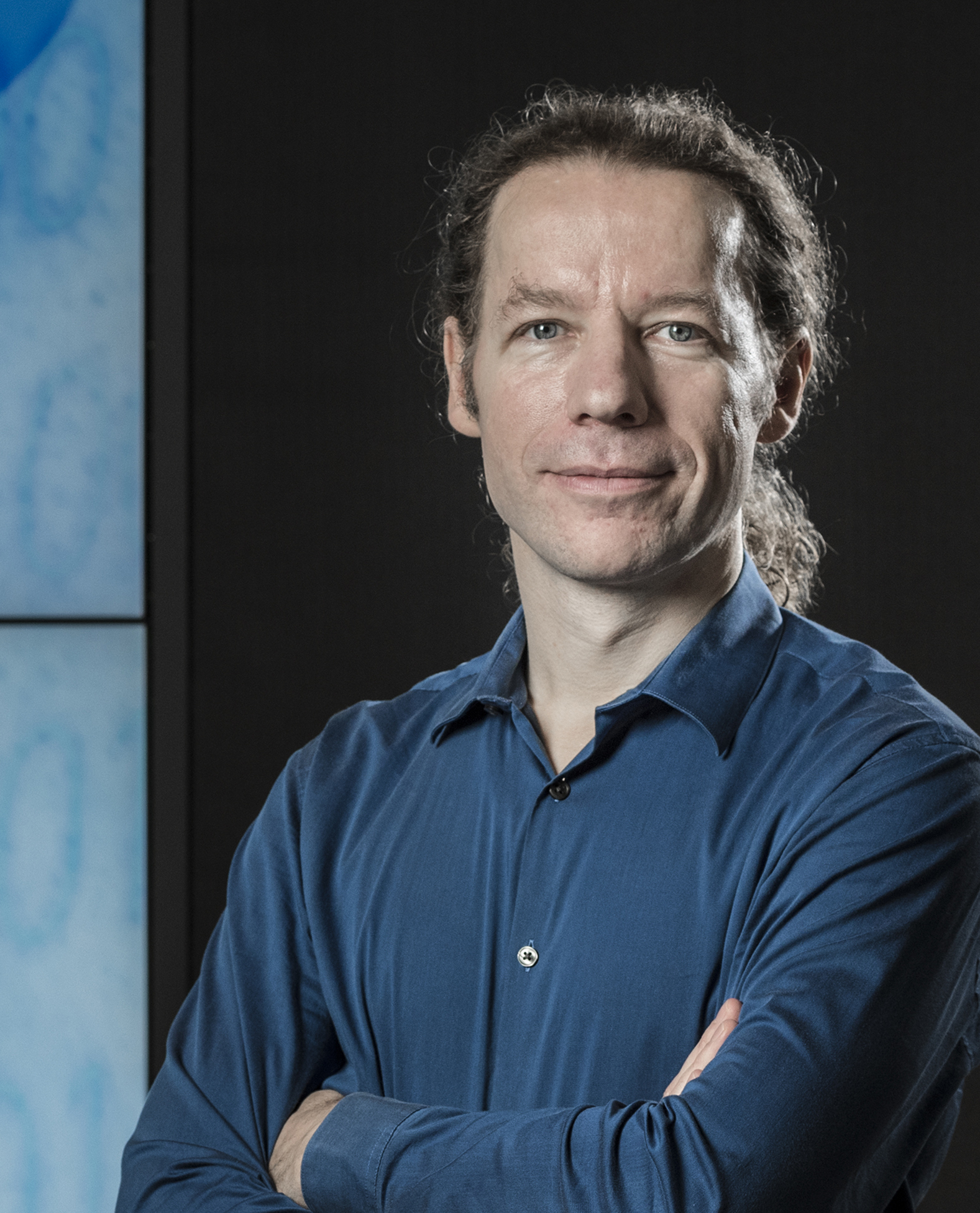}}]{Gunnar R\"atsch} leads the Biomedical Informatics group at the Institute of Machine Learning in the Computer Science Department at ETH Zurich and is an Adjunct Faculty at the University Hospital Zurich (USZ). In 2002, he received the Michelson award for his Ph.D. work with in Machine Learning and in 2007 he was awarded the Olympus prize from the German Association for Pattern Recognition for his work on Boosting. He has extensive research expertise in Genomics and Machine Learning and has published more than 150 peer-reviewed publications with >40,000 citations (Google Scholar) with an H-index of 72. His group has developed advanced algorithms for large-scale learning and data integration, automated genome annotation, and machine learning in medicine. A hallmark of his group is tackling highly relevant biomedical problems that require non-trivial technical solutions. Current research focuses include i) novel machine learning techniques for multi-modal data integration, ii) statistical genetics approaches to elucidate gene regulation, and iii) understanding transcriptome and proteome plasticity.
\end{IEEEbiography}

\EOD

\end{document}